\documentclass[runningheads]{llncs}
\usepackage{esvect}
\usepackage[T1]{fontenc}
\usepackage{graphicx,verbatim}
\usepackage{orcidlink}
\usepackage{amsmath}
\usepackage{multirow} 
\usepackage{subcaption} 
\usepackage[utf8]{inputenc}
\usepackage{pifont}
\begin{document}

\title{Diffusion-Based Tumor Inpainting for Renal Segmentation under Clinical Data Scarcity}
\titlerunning{DDPM Augmentation for Renal Tumor Segmentation}



\author{
Ekaterina Sedykh\inst{1,2,3}\orcidlink{0000-0002-0972-3008} \and
Salme Ussanov\inst{1}\orcidlink{0009-0001-7250-7746} \and
Dmytro Fedorenko\inst{1,4,5}\orcidlink{0000-0001-7883-7001} \and
Dmytro Fishman\inst{1,6,7}\orcidlink{0000-0002-4644-8893}
}

\authorrunning{E. Sedykh et al.}
\institute{
Institute of Computer Science, University of Tartu, Tartu, Estonia\\
\and
Institute of Tissue Medicine and Pathology, University of Bern, Bern, Switzerland\\
\and
Department of Digital Medicine, University of Bern, Bern, Switzerland\\
\and
School of Computer Science and Centre for Intelligent Machines, McGill University, Montréal, Canada\\
\and
Mila – Québec Artificial Intelligence Institute, Montréal, Canada\\
\and
Better Medicine OÜ, Tartu, Estonia\\
\and
STACC OÜ, Tartu, Estonia\\
\email{ekaterina.sedykh@unibe.ch, dmytro.fishman@ut.ee}
}

\maketitle              
\begin{abstract}
Deep learning segmentation of renal tumors requires large annotated datasets, yet clinical deployments typically offer only a handful of tumor-positive cases from the target site. We propose a diffusion-based inpainting framework that synthesizes anatomically plausible renal tumors within healthy CT scans, requiring no additional annotation, and provide the first systematic comparison of 2D, 2.5D, and full 3D (MAISI) synthesis strategies for this task. Training the diffusion model on public data (KiTS23, KIRC) and evaluating nnU-Net segmentation on a internal cohort across three low-data regimes, we find that 2.5D and 3D augmentation substantially reduce false positives (from $\sim$18--20\% to $\sim$3--6\%) while maintaining Dice, whereas 2D provides no consistent benefit. Crucially, the proposed 2.5D method matches full 3D synthesis on every metric at substantially lower computational cost, indicating that local volumetric consistency alone is sufficient for effective augmentation in data- and resource-scarce
clinical settings.

\keywords{Diffusion Models  \and Synthetic Data Generation \and CT Scans \and Kidney Tumor.}

\end{abstract}
\section{Introduction}
Medical image segmentation - the delineation of anatomical structures and pathological regions such as organs and lesions - is a fundamental step in medical image analysis, supporting applications including computer-aided diagnosis, surgical planning, and treatment monitoring \cite{lepcha2025deep}. For renal cancer in particular, accurate tumor segmentation on CT supports diagnosis, surgical planning, and follow-up, but performing it manually is time-consuming and subject to inter-observer variability, motivating reliable automated methods \cite{heller2021state}.

Deep learning-based segmentation methods have become the dominant approach to automating this task, but they require substantial annotated datasets, which are often limited in medical imaging due to privacy concerns, annotation costs, and data scarcity \cite{kebaili2023deep,wang2024comprehensive}. This limitation is especially pronounced for tumor-positive cases, which are considerably rarer than healthy examples in most clinical archives, producing a class imbalance that further constrains supervised training \cite{kebaili2023deep}. These constraints are most acute in realistic deployment scenarios, such as collaborations between research laboratories or industry partners and clinical institutions. Even when a hospital holds extensive imaging archives, expert annotations are costly and typically available for only a small subset of cases \cite{wang2024comprehensive}. As a result, models must frequently be developed using only a minimal amount of tumor-positive data obtained from the target clinical site, while still generalizing reliably to that institution's imaging protocols and patient population.

Several strategies have been proposed to alleviate this data limitation. Classical data augmentation (e.g. spatial and intensity transformations) enlarges datasets cheaply, but cannot introduce genuinely new pathology and often produces limited, unconvincing variations \cite{kebaili2023deep}. Semi-supervised and weakly supervised learning reduce annotation demands but still rely on a representative labeled core. An alternative is to synthesize additional training data directly - by generating new tumor-positive examples, synthetic data generation is a promising approach to augment existing datasets and improve model generalization, provided the generated data is sufficiently realistic and diverse. More recently, diffusion models have shown great promise due to their stable training dynamics and their ability to generate highly realistic and diverse data \cite{ho2020denoising}. Whether this promise translates into measurable downstream segmentation gains under realistic, data-scarce clinical conditions, however, remains an open question.

In this work, we propose a diffusion-based data augmentation framework for renal tumor segmentation that synthesizes anatomically plausible kidney tumors directly within healthy CT scans, investigating their impact on downstream renal tumor segmentation performance. Our goal is to demonstrate how a segmentation model can be effectively built under such constraints, using only limited annotated data from a clinical partner and synthetic augmentation to compensate for tumor scarcity.

\begin{figure}[t]
    \centering
    \includegraphics[width=\textwidth]{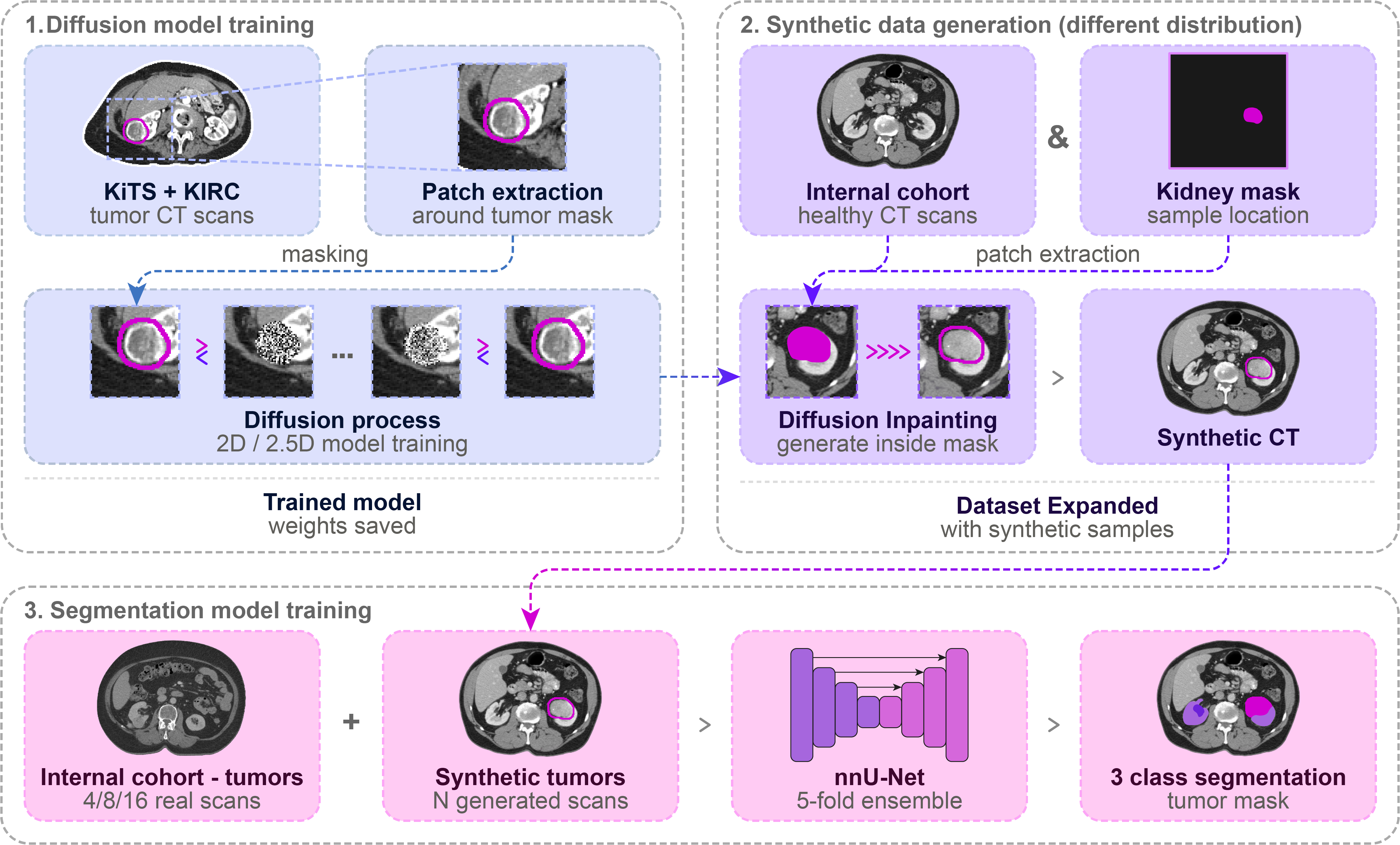}
    \caption{The complete framework operates in three stages. (1) A DDPM is trained on public tumor-containing CT datasets (KiTS23, KIRC) with soft-tissue windowing and normalization. Both 2D and 2.5D variants are implemented. (2) For synthesis, candidate tumor masks from KiTS are geometrically processed and placed into tumor-free control scans from internal cohort using an anatomically-constrained placement module. Hard constraints prevent anatomically impossible placements (bone overlap, non-body regions), while soft penalties encourage realistic kidney-tumor interactions. The DDPM then performs context-aware inpainting to synthesize realistic tumor texture. (3) nnU-Net is trained under various data regimes and evaluated on held-out test sets using Dice coefficient and false positive analysis.}
    \label{fig:pipeline_overview}
\end{figure}

\section{Related Work}
Data scarcity and class imbalance pose significant challenges for medical image segmentation, especially in oncology applications, where tumor-positive cases are typically limited as malignant findings are inherently less frequent than healthy examples, and expert voxel-level annotation of those cases is costly and time-consuming, so labeled tumor data accumulates slowly even at large centers \cite{kebaili2023deep,wang2024comprehensive}. To address these issues, synthetic data generation strategies have become an active area of research. Early augmentation methods take the route of augmenting existing data, applying classical image-processing operations such as elastic transformations and intensity variations to existing scans. Because every augmented sample is a transformed copy of a real one, these methods cannot introduce new pathology or add genuinely novel tumor appearances, and aggressive transformations often produce unrealistic, anatomically implausible variations, limiting their effectiveness in complex clinical scenarios \cite{nalepa2019data,goceri2023medical}. Generative models, by contrast, take the route of producing genuinely new data: they learn the underlying data distribution and can synthesize new examples, making them better suited to expanding the effective pool of tumor-positive cases.

Deep generative models, such as Generative Adversarial Networks (GANs) - in which a generator and a discriminator are trained against one another, the generator learning to produce images the discriminator cannot distinguish from real ones - have demonstrated substantial promise in creating realistic synthetic medical images \cite{yi2019generative}. For example, GAN-based approaches were successfully employed in brain tumor synthesis, significantly improving segmentation model robustness \cite{han2019combining,shin2018medical,foroozandeh2020synthesizing}. Nonetheless, GANs are known for experiencing unstable trainings, mode collapse, and difficulties in scaling to volumetric data \cite{saad2024survey}. Variational Autoencoders (VAEs)~\cite{kingma2013auto} provide an alternative generative approach that, rather than relying on adversarial training, learn to encode images into a probabilistic latent space and reconstruct them, optimizing a likelihood-based objective. This gives them more stable training than GANs, and they have been applied to medical image synthesis tasks such as augmenting CT and MRI training sets for segmentation~\cite{rais2024exploring}. However, VAEs tend to produce blurry outputs due to pixel-wise reconstruction losses, limiting their realism and clinical applicability.

More recently, Denoising Diffusion Probabilistic Models (DDPMs) have emerged as an advanced class of generative models, capable of generating high-quality images with superior diversity and detail \cite{ho2020denoising,luo2025review}. DDPMs have already demonstrated strong results in various medical imaging tasks, including synthetic chest X-ray generation \cite{huijben2024denoising}, MRI brain lesion synthesis \cite{dorjsembe2024conditional}, and pathology slide generation \cite{moghadam2023morphology}. Due to their stable training dynamics and high-fidelity outputs, DDPMs hold significant promise for medical data augmentation.

We are interested specifically in using such generative models to synthesize additional tumor-positive training data for renal tumor segmentation, thereby compensating for the scarcity of annotated tumor cases. While segmentation of the kidney and renal tumors on CT is itself a well-studied task, the use of generative models and diffusion models in particular to augment training data for it has received comparatively little attention. Prior studies have focused on traditional data augmentation or GAN-based synthetic tumor generation \cite{jin2021free}, and, more recently, on diffusion-based multi-organ tumor synthesis in CT \cite{chen2024towards}, yet diffusion-driven renal tumor synthesis as an augmentation technique has not been explored in depth.  A key design choice in any such approach is whether to generate images unconditionally or through inpainting. Generating full synthetic scans unconditionally requires the model to reproduce the entire abdominal anatomy from scratch, which is difficult to do plausibly, offers no control over where or whether a tumor appears, and produces images without reliable, pre-existing segmentation labels, which is critical for augmentation usage. Inpainting avoids these problems, rather than generating full synthetic scans unconditionally, it inserts pathology into existing healthy images, preserving global anatomy while locally modifying only the pathological region. Because the surrounding anatomy is real and the inserted lesion is placed at a known location, the resulting images are both anatomically consistent and automatically annotated, making inpainting a particularly promising direction for augmentation.

Recent diffusion-based inpainting methods have demonstrated strong performance in free-form region completion and context-aware image reconstruction, benefiting from explicit conditioning on surrounding pixels and iterative denoising dynamics \cite{lugmayr2022repaint,song2020score}. In the medical domain, inpainting approaches have been applied to lesion removal, anomaly modeling, and anatomy-aware reconstruction tasks \cite{wolleb2022diffusion,pinaya2022brain}. These properties make diffusion-based inpainting especially suitable for tumor synthesis, where preserving global anatomy while locally modifying pathological regions is essential.

Despite these advances, the application of diffusion-based inpainting to renal tumor synthesis as a training augmentation strategy remains largely unexplored. Existing work on synthetic tumor generation has predominantly focused on GAN-based approaches \cite{jin2021free} or non-tumor specific unconditional volumetric synthesis, and has rarely been evaluated in strictly limited-data regimes representative of real clinical deployment conditions. 

The extent to which synthetic augmentation can compensate for a scarcity of real tumor-positive cases in this setting remains underexplored, and few studies evaluate synthetic tumor generation under the strictly limited-data conditions representative of real clinical deployment. Moreover, to our knowledge, 2D, 2.5D, and full 3D diffusion-based synthesis strategies have not been directly compared for this task under matched conditions. These strategies differ in how much volumetric context they use during generation: 2D synthesizes each slice independently, 2.5D conditions on a small number of neighbouring slices to encourage continuity in the through-plane direction, and full 3D generates the entire volume jointly. This choice trades off computational cost against volumetric coherence, yet it remains unclear whether the coherence afforded by higher-dimensional generation is necessary to achieve downstream segmentation gains, or whether limited slice context suffices.

This work addresses these gaps directly. We propose a diffusion-based inpainting framework that synthesizes anatomically plausible renal tumors within healthy CT scans from a clinical partner hospital, requiring no additional annotation of the generated data. We evaluate the impact of synthetic augmentation on nnU-Net~\cite{isensee2021nnu} segmentation performance across three low-data regimes (2.5\%, 5\%, and 10\% of available tumor cases), and provide the first systematic comparison of 2D, 2.5D, and 3D synthesis strategies under matched experimental conditions.

\section{Methods}
\subsection{Data overview}
We utilize two publicly available datasets: Kidney Tumor Segmentation Challenge 2023 (KiTS23) dataset \cite{heller2023kits21} and The Cancer Genome Atlas Kidney Renal Clear Cell Carcinoma (TCGA-KIRC, simplified to KIRC) dataset \cite{Akin2016_TCGA_KIRC}, and a internal clinical dataset. The internal cohort from Tartu University Hospital (TUH) contains 191 tumor-positive cases and 200 healthy control cases, where controls show no radiological evidence of malignant renal lesions, but can contain cysts. Each TUH dataset scan was manually segmented by clinical specialists, providing kidney and tumor annotations used for downstream segmentation and validation. The diffusion model was trained exclusively on publicly available tumor datasets (KiTS23 and KIRC), while the internal cohort was used solely for anatomically constrained tumor placement, synthetic tumor generation in healthy controls, and downstream segmentation experiments.

As diffusion models operate on normalized inputs in the range $[-1, 1]$, intensities were windowed to the soft-tissue HU range ([-200, 300] HU) prior to normalization. The DDPM was trained and applied in this windowed domain, and synthetic results were mapped back to the same windowed range. To ensure comparability, nnU-Net was also trained and evaluated on the same windowed data rather than raw HU values. This choice is consistent with clinical practice, where soft-tissue windowing is typically used to improve kidney and tumor visualization.

\subsection{DDPM}
We implement a Denoising Diffusion Probabilistic Model based on the Palette framework \cite{saharia2022palette}, an image-to-image diffusion approach originally designed for conditional image generation tasks. The model architecture follows a U-Net backbone with attention layers, which enhances details and contextual consistency. We took the public implementation\footnote{\url{https://github.com/Janspiry/Palette-Image-to-Image-Diffusion-Models}} as a starting point and adapted it to our datasets and to the tumor inpainting task by modifying the data-processing pipeline to handle single 2D slice and 2.5D stacks

We operate directly in image space rather than in latent space. While latent diffusion models (LDMs) \cite{rombach2022high} offer computational efficiency through autoencoder compression, this compression can discard fine structural and intensity details that are critical in radiological data, in particular, subtle Hounsfield Unit variations and texture continuity. Operating in image space avoids this bottleneck and allows the model to learn the true data distribution directly.

Training was performed on tumor-containing slices extracted from the KiTS23 and KIRC datasets, with no exposure to internal TUH data during diffusion model optimization. This cross-dataset setup means the model is applied at inference time to a different data distribution, which we consider a realistic constraint for clinical deployment scenarios where target-site annotated data is scarce.

\subsubsection{2D Method}
In the 2D approach, individual slices are treated independently. Each slice containing the tumor mask is processed separately, and synthetic tumor textures are generated independently for each slice. This method provides lightweight training, fast iterations, and abundant slice-wise examples but lacks volumetric context, which can lead to artifacts and inconsistent tumor appearances across slices.

Concretely, generation is confined to the region defined by a binary tumor mask $m$ (with $m=1$ inside the tumor), while the surrounding anatomy is held fixed. During training, noise is added only within the mask, leaving the known region clean,
\begin{equation}
\tilde{y}_\gamma = m \odot y_\gamma + (1-m)\odot y_0,
\end{equation}
where $y_0$ is the clean slice and $y_\gamma$ its noised version at noise level $\gamma$. The U-Net receives the conditioning slice and this composited input concatenated along the channel dimension, and the loss is applied only inside the mask:
\begin{equation}
\mathcal{L} = \big\lVert\, m\odot\big(\epsilon - f_\theta([\,y_{\text{cond}},\,\tilde{y}_\gamma\,], \gamma)\big) \big\rVert_{1}.
\end{equation}
At inference, after every reverse diffusion step, the region outside the mask is reset to the original slice and only the generated content inside the mask is retained,
\begin{equation}
y_{t-1} \leftarrow (1-m)\odot y_0 + m\odot y_{t-1},
\end{equation}
so that only the tumor region is synthesized while the host anatomy is preserved exactly. In this 2D configuration, the network takes 2 input channels (conditioning slice and noisy target) and produces a single output channel. This makes training lightweight and inference fast, and provides abundant slice-wise training examples, but it gives the model no volumetric context, which can lead to artifacts and inconsistent tumor appearances across adjacent slices. Generating a synthetic tumor for one slice peaks at roughly 5~GB of GPU memory.

\subsubsection{2.5D Method}
The 2.5D approach uses the same conditional inpainting formulation as the 2D model, but considering auxiliary adjacent slices, capturing limited 3D contextual information by stacking neighbouring slices as input. Specifically, we provide two preceding slices while generating a target slice, so that the synthesized tumor is coherent with the anatomy already present below it. The tumor mask is applied to the target slice, the two auxiliary slices are provided as unmasked context. Accordingly, the network takes 6 input channels - the conditioning and noisy target for the three stacked slices - and produces 3 output channels, in place of the 2 and 1 channels of the 2D model. Generation proceeds as a rolling window along the $z$-axis - each newly generated slice becomes context for the next (Figure~\ref{fig:synthesis_comparison}). The tumor volume is synthesized slice by slice in order, and each newly generated slice is carried forward to serve as an auxiliary context slice for the next step. In this way the model conditions not on the original healthy slices but on its own previously generated tumor content, propagating a consistent tumor appearance through the volume and enforcing continuity of texture and boundary in the through-plane direction. This autoregressive scheme is the key difference from the 2D model: it introduces volumetric coherence at generation time while retaining the lightweight, slice-based architecture, and therefore incurs only a modest increase in computational cost over the 2D approach. Its main limitation is the restricted through-plane receptive field: because only two neighbouring slices are seen at a time, consistency is enforced locally rather than globally, which may permit gradual drift for large
lesions spanning many slices.

\subsubsection{3D Method (MAISI)}
As a third method, we employ MAISI (Medical AI for Synthetic Imaging)~\cite{guo2025maisi}, a large-scale 3D latent diffusion model trained on over 10,000 CT volumes from diverse anatomical regions. MAISI comprises three components: a VAE-GAN that encodes CT images into a compressed latent representation, a diffusion model that acts as the main generative component, and a ControlNet that enables task-specific conditioning. Operating on full volumetric inputs, it preserves spatial consistency across the synthesized scan, with generation conditioned on an organ segmentation mask covering 127 anatomical structures. Kidney tumor is not among these 127 structures, so we fine-tune MAISI specifically for kidney tumor generation on the same KiTS23 and KIRC datasets. Each scan was resampled to $512 \times 512 \times 256$ voxels at 1~mm slice thickness in RAS orientation. 

As MAISI was designed to generate whole volumes, we adapt it for inpainting. Rather than initializing pure noise across the entire volume, we encode a non-tumor CT scan into the latent space using MAISI's VAE encoder and initialize noise only within the tumor region defined by the tumor mask. This latent is passed to the ControlNet and diffusion model, and after each diffusion timestep we reinstate the original latent everywhere except inside the tumor region, updating only the tumor features with the diffusion output. After the final timestep, the latent is decoded back to image space by the VAE decoder. This confines synthesis to the tumor and its immediate surroundings while leaving the surrounding anatomy unchanged.

Although the latent representation makes 3D diffusion require less memory than the full image space, MAISI remains substantially more resource-intensive than the slice-based approaches. Generating a single volume peaks at approximately 60~GB of allocated GPU memory; the encoding step alone, performed with overlapping patches to limit memory use, peaks between 26 and 139~GB depending on the number of patch splits. By contrast, our slice-based methods operates at a small fraction of this cost. We therefore use MAISI as a strong 3D baseline to test whether full volumetric synthesis offers measurable benefits over our proposed 2.5D approach.

\begin{figure}[htbp]
    \centering
    \includegraphics[width=\linewidth]{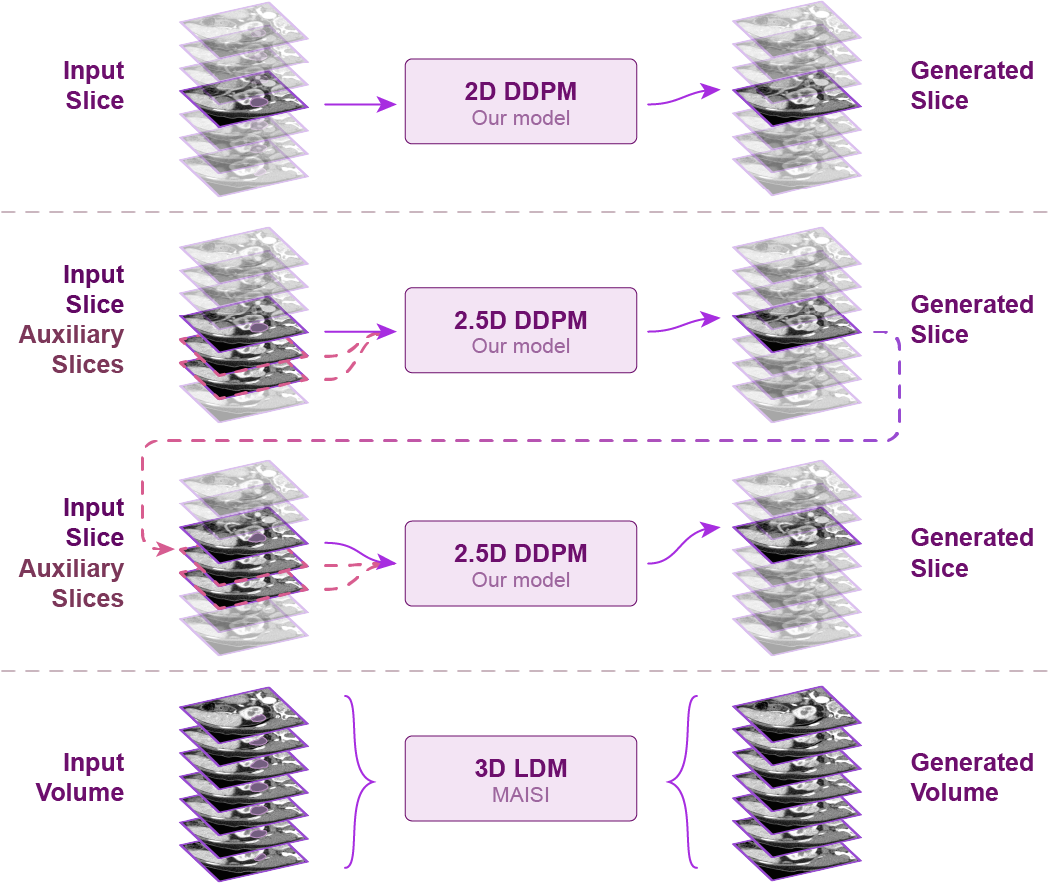}
    \caption{Comparison of the three synthetic tumor generation strategies.
    \textbf{2D (ours):} a single target slice is inpainted independently inside the tumor mask, with no cross-slice context.
    \textbf{2.5D (ours):} the target slice is generated using the two preceding slices as context (without mask); the generated slice then feeds into the next iteration as context, enabling a rolling window over the volume.
    \textbf{3D (MAISI):} the full CT volume is synthesized jointly in a single pass through a latent diffusion model.}
    \label{fig:synthesis_comparison}
\end{figure}

The three synthesis strategies are illustrated in Figure~\ref{fig:synthesis_comparison}.

\subsection{Anatomically-Constrained Synthetic Tumor Placement} \label{sec:placement}
Tumor placement was performed on TUH dataset control scans only, enabling the synthesis of tumor-positive cases while preserving original scan geometry and acquisition characteristics. To ensure that synthesized lesions are spatially and anatomically realistic, we developed an automated tumor placement module operating on resampled 1\,mm isotropic CT volumes. The approach integrates geometric kidney surface extraction, multi-organ exclusion rules, and a biologically motivated scoring model that prioritizes plausible interaction between generated tumors and renal parenchyma.

\subsubsection{Preprocessing and resampling.}
For each scan, TotalSegmentator \cite{wasserthal2023totalsegmentator} was used to obtain body, bone, and abdominal organ masks. Kidney annotations (label 1) from our dataset were retained, whereas cystic regions (label 3) were explicitly excluded to avoid pathological ambiguity. All masks and the native CT volume were resampled to 1\,mm$^{3}$ resolution, enforcing spatial consistency across datasets.

Candidate tumors were drawn from the KiTS dataset's masks. Each tumor underwent morphological cleaning and surface smoothing to remove high-frequency artifacts and improve geometric realism. When placed, each lesion was cropped to its minimal bounding box and uniformly scaled so that its final volume occupied between 5--35\% of the recipient kidney volume, preventing implausibly small or excessively large synthetic lesions.


\begin{figure}[ht]
  \centering
  \begin{subfigure}[b]{0.3\linewidth}
    \includegraphics[width=\linewidth]{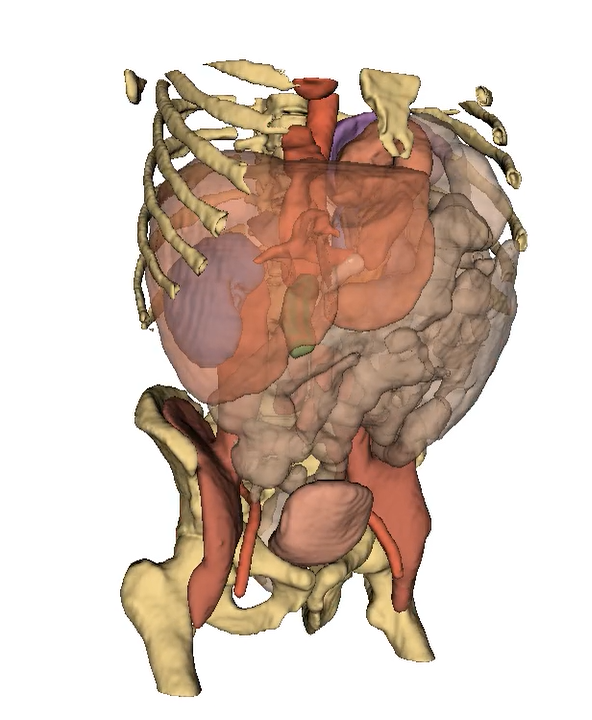}
    \caption{}
    \label{fig:placement_a}
  \end{subfigure}
  \hfill
  \begin{subfigure}[b]{0.6\linewidth}
    \includegraphics[width=\linewidth]{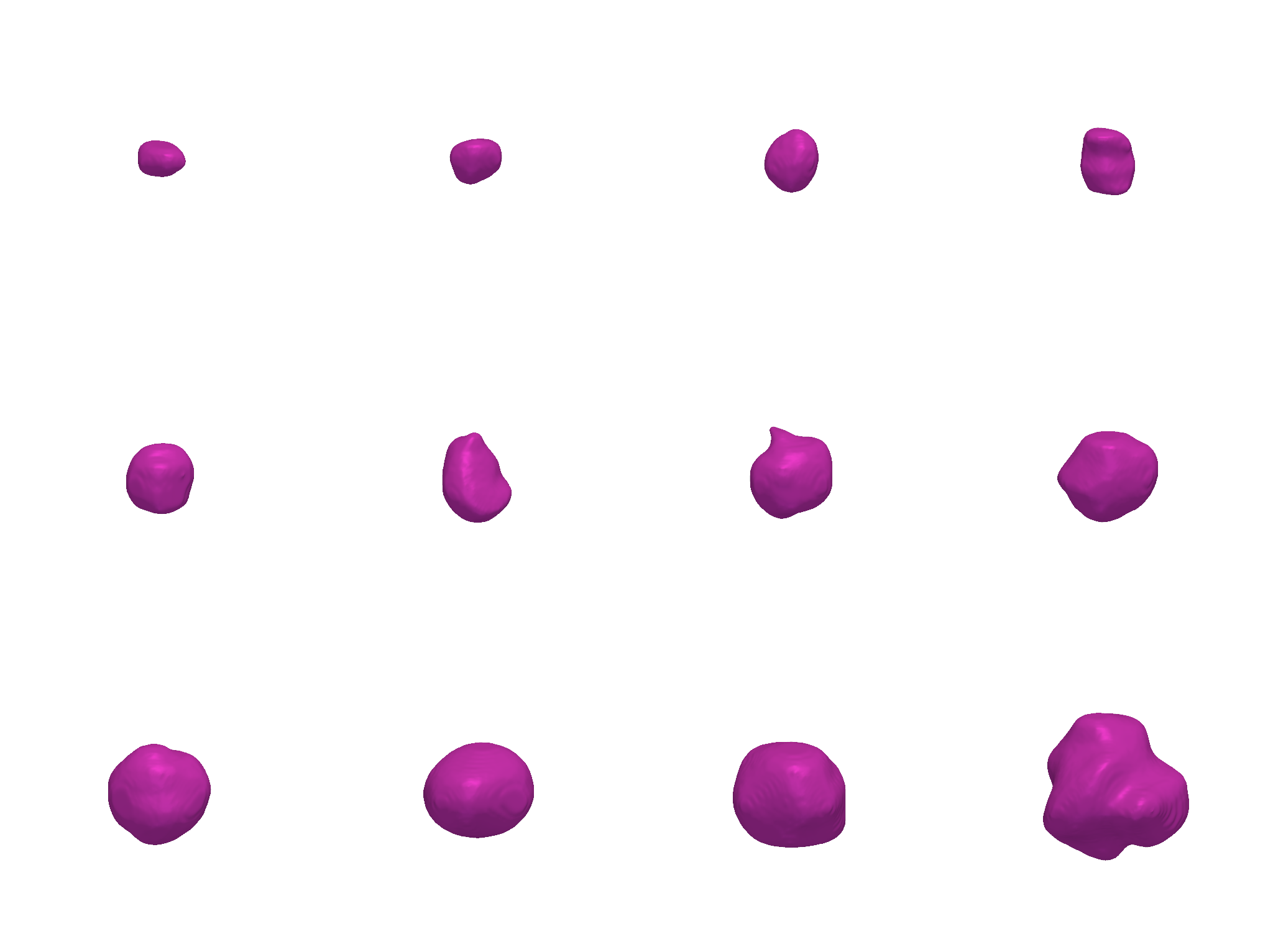}
    \caption{}
    \label{fig:placement_b}
  \end{subfigure}
  \caption{Inputs to the anatomically-constrained placement pipeline.
  \textbf{(a)} Multi-organ segmentation of a host control scan obtained with TotalSegmentator, providing the bone, organ, and vascular  structures used as hard and soft placement constraints.
  \textbf{(b)} A representative subset of the candidate tumor library: real tumor meshes extracted from KiTS cases, from which lesions are sampled, scaled, and inserted. Tumors are shown at a common scale, illustrating the diversity of size and shape available for synthesis.}
  \label{fig:placement_inputs}
\end{figure}

\subsubsection{Surface-anchored placement.}
Synthetic tumors were positioned by first sampling a random kidney surface voxel and attempting to centre the lesion around it. If insufficient kidney-tumor contact was achieved, the lesion was iteratively shifted inward along an approximate surface normal direction, or moved with small displacements, until a valid attachment was formed. This enforces mandatory physical contact between the tumor and kidney surface.

\subsubsection{Anatomical scoring model.}
Each candidate placement was evaluated using a penalty-based scoring objective with two components. Hard rejection rules were applied first: the tumor must intersect the kidney surface, no voxels may lie outside the segmented body envelope, no overlap with bone structures is permitted, and the tumor must not intrude into cystic regions. Placements violating any of these conditions were rejected immediately without further scoring.

Remaining candidates were then scored using soft anatomical penalties. A contact-regularization term encouraged approximately 30\% kidney--tumor overlap:
\[
    P_{\text{contact}} = 10 \cdot (\text{contact} - 0.3)^2
\]
An organ-intrusion penalty was computed from tumor overlap fractions weighted by anatomical relevance: soft organs (e.g.\ liver, bowel, pancreas) were assigned weight~50, while major vasculature (aorta, vena cava, portal vein, iliac vessels) received higher weights of~200 to reflect the greater clinical implausibility of vascular intrusion. The total placement score was the sum of soft penalties, with lower scores indicating more plausible insertion sites.

\subsubsection{Multi-candidate optimization and rejection sampling.}
For each tumor mask, 30 candidate placements were sampled and scored. The lowest-penalty configuration was selected; if its score exceeded a plausibility threshold $\tau = 0.3$, the tumor was rejected entirely and a new lesion was drawn from the library. This iterative rejection sampling step suppresses anatomically inconsistent insertions. Because the individual weights are meaningful only relative to $\tau$ and to one another, we chose them so that each term encodes an interpretable tolerance on the corresponding anatomical violation.

\subsection{Evaluation metrics}
We used the Fréchet Inception Distance (FID) to quantify realism and distributional similarity between real and synthetic tumors \cite{heusel2017gans}. Since direct FID on full CT scans would be dominated by background tissue, we computed FID only on cropped tumor regions. FID was used as a supplementary measure to assess local texture realism of synthetic tumors, while segmentation performance served as the primary criterion for evaluating the effectiveness of synthetic augmentation.

The nnU-Net~\cite{isensee2021nnu} was trained using the \texttt{3d\_fullres} configuration on different combinations of real and synthetic data. Dice score was reported separately for tumor, cyst, and kidney classes, with main focus on tumor score improvement. The fraction of control scans with any predicted tumor voxels is reported as false-positive rate, and additionally number of incorrect voxels in all scans is included.

Different training set compositions and tumor-positive data regimes were explored to assess the impact of synthetic augmentation, as detailed in the Results section.

\section{Results and Discussion}
We evaluate the effectiveness of diffusion-based synthetic tumor augmentation for renal tumor segmentation under data-scarce training regimes. All experiments are conducted on a held-out TUH test set comprising 38 tumor-positive and 40 control scans, enabling direct assessment of segmentation quality and specificity in a realistic clinical setting. We compare three synthetic augmentation strategies - 2D, 2.5D (ours), and 3D (MAISI) - against a real-data-only baseline across three low-data regimes.

\subsection{Experimental Setup}
To simulate data-scarce clinical scenarios, we considered training subsets containing 2.5\%, 5\%, and 10\% of the available internal TUH cohort tumor-positive training scans, corresponding to 4, 8, and 16 scans respectively. For each method, synthetic tumors were generated by inserting a single diffusion-generated tumor into each TUH dataset control training scan, yielding 160 synthetic tumor-positive cases. Ten random subsets were sampled from real data per regime to assess the sensitivity to dataset composition, and performance is reported as mean $\pm$ standard deviation across these splits. All models use 5-fold nnU-Net ensemble training with postprocessing filtering applied uniformly across conditions.

We evaluated four training configurations per regime: (i) real data only, (ii) real + 2D synthetic, (iii) real + 2.5D synthetic, and (iv) real + 3D synthetic (MAISI).

\subsection{Extreme Low-Data Regime: 2.5\% Real Training Data (4 Scans)}

At the most constrained setting, real-only training already achieved a mean tumor dice of $0.844 $ (Table~\ref{tab:results_2_5}). Adding 2D synthetic augmentation yielded no improvement, with Dice dropping to $0.774$, suggesting that 2D-generated tumors introduce features that do not transfer well to volumetric segmentation at such limited real-data levels. In contrast, 2.5D and 3D augmentation left Dice statistically unchanged relative to real-only ($0.855$ and $0.819$). 

\begin{figure}[ht]
    \centering
    \includegraphics[width=\linewidth]{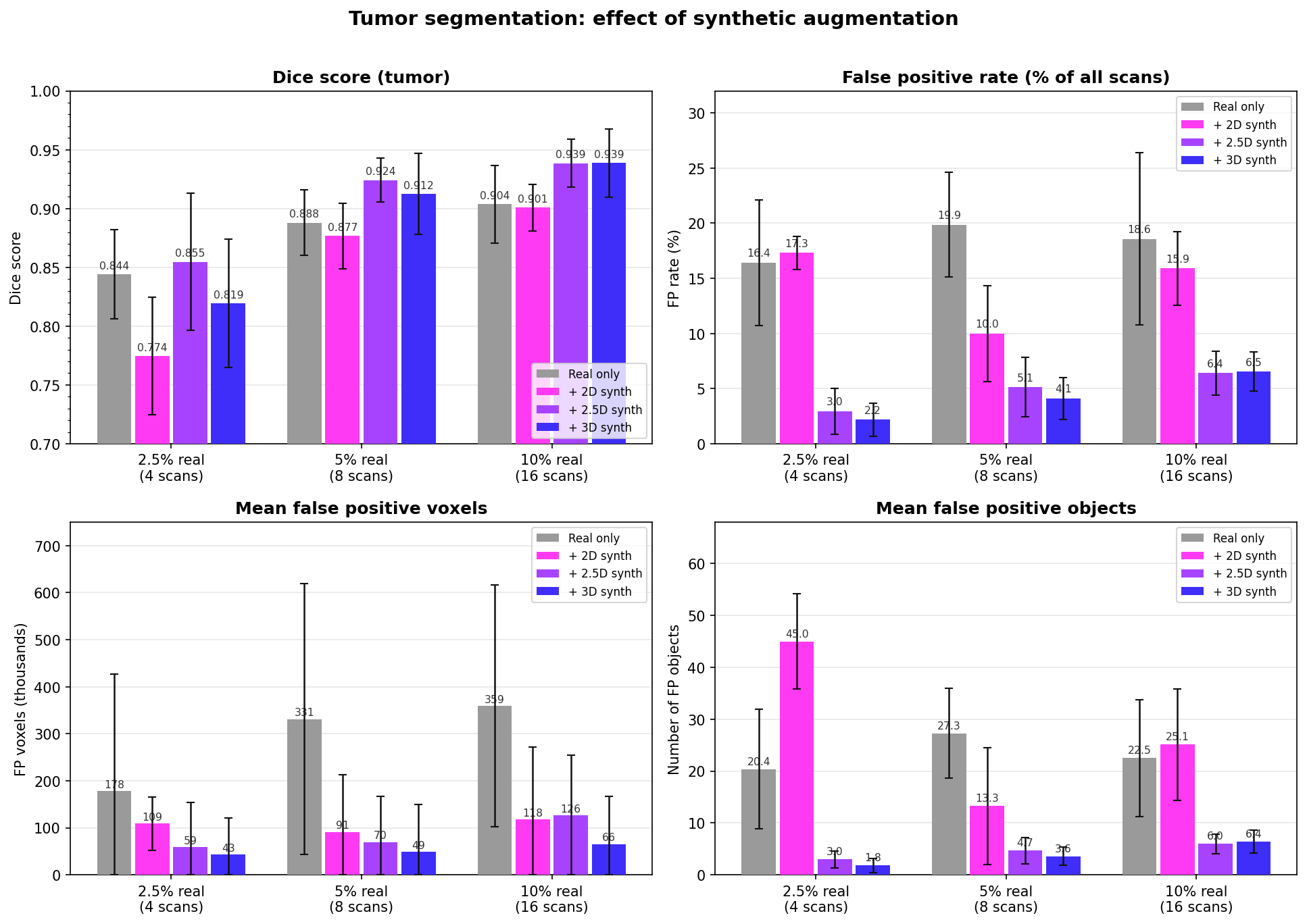}
    \caption{Segmentation performance across real-data regimes and synthetic augmentation strategies. Mean Dice score (tumor), false positive rate (\% of all scans), mean FP voxels, and mean FP objects, averaged over 10 random splits with 5-fold nnU-Net ensembles. Error bars show standard deviation. FP voxel error bars are clipped at zero.}
    \label{fig:barplots}
\end{figure}

Real-only and 2D-augmented models both produced high false positive rates ($16.4\%$ and $17.3\%$), whereas 2.5D and 3D augmentation suppressed false positives dramatically, cutting the FP rate to roughly $2$--$3\%$ and reducing false positive objects by close to an order of magnitude (Table~\ref{tab:results_2_5}). This false-positive suppression, both in the occurrence rate and also number of FP voxels, is the main benefit of synthetic augmentation at this regime.

Substantial variability was observed across the ten 2.5\% subsets: dice ranged from $0.733$ to $0.923$ under 2.5D augmentation (range 0.19), and FP voxel counts varied by over an order of magnitude. This reflects sensitivity to the selection of which four scans rather than optimization instability. Notably, the subset-induced variation exceeded the mean Dice difference between any two augmentation conditions, indicating that at 4 training scans, sample selection dominates model behavior.

\begin{table}[t]
\centering
\caption{Segmentation performance at 2.5\% real training data (4 scans), averaged over 10 random splits with 5-fold ensembles. Bold denotes best per metric.}
\label{tab:results_2_5}
\begin{tabular}{lcccccc}
\hline
Condition & Dice & IoU & FP rate (\%) & FP objects & FP voxels \\
\hline
Real only      & $0.844 \pm 0.038$ & $0.732 \pm 0.058$ & $16.4$ & $20.4 \pm 11.5$ & $177{,}750$ \\
+ 2D synth     & $0.774 \pm 0.050$ & $0.634 \pm 0.067$ & $17.3$ & $45.0 \pm 9.1$  & $108{,}829$ \\
+ 2.5D synth   & $\mathbf{0.855 \pm 0.058}$ & $\mathbf{0.750 \pm 0.085}$ & $2.95$ & $3.0 \pm 1.6$ & $58{,}848$ \\
+ 3D synth     & $0.819 \pm 0.055$ & $0.697 \pm 0.076$ & $\mathbf{2.18}$ & $\mathbf{1.8 \pm 1.3}$   & $\mathbf{42{,}709}$ \\
\hline
\end{tabular}
\end{table}

\subsection{Moderate Low-Data Regimes: 5\% and 10\% Real Training Data}

As real data increased, the real-only baseline improved significantly from 2.5\% to 5\% (Dice $0.844 \to 0.888$, $p = 0.018$), but not from 5\% to 10\% ($p = 0.29$), indicating diminishing returns beyond eight real scans in this task (Table~\ref{tab:results_all}).

At 5\%, 2.5D and 3D augmentation both significantly outperformed real-only on Dice ($0.924$ and $0.912$ vs.\ $0.888$), while 2D again gave no significant benefit ($0.877$). At 10\%, all conditions converged toward similar Dice scores ($0.901$--$0.939$), with differences between 2.5D, 3D, and real-only no longer significant.

The FP suppression advantage of 2.5D and 3D augmentation persisted across all regimes. Both methods held the FP rate to roughly $4$--$6\%$ across the 5\% and 10\% regimes, compared with $18$--$20\%$ for real-only, with correspondingly large reductions in FP object counts (Table~\ref{tab:results_all}). 2D augmentation was again the outlier: its FP behavior was unstable, producing far more FP objects than any other condition at 2.5\% before partially normalizing at higher regimes. Across all three regimes, then, 2.5D and 3D consistently improve specificity while matching or exceeding real-only Dice, whereas 2D does neither reliable.

\begin{table}[t]
\centering
\caption{Mean Dice and FP rate across all regimes and augmentation conditions (10 splits, 5-fold ensembles).}
\label{tab:results_all}
\begin{tabular}{llccc}
\hline
Condition & Metric & 2.5\% (4 scans) & 5\% (8 scans) & 10\% (16 scans) \\
\hline
\multirow{2}{*}{Real only}    & Dice    & $0.844 \pm 0.038$ & $0.888 \pm 0.028$ & $0.904 \pm 0.033$ \\
                               & FP rate & $16.4\%$          & $19.9\%$          & $18.6\%$ \\
\hline
\multirow{2}{*}{+ 2D synth}   & Dice    & $0.774 \pm 0.050$ & $0.877 \pm 0.028$ & $0.901 \pm 0.020$ \\
                               & FP rate & $17.3\%$          & $10.0\%$          & $15.9\%$ \\
\hline
\multirow{2}{*}{+ 2.5D synth} & Dice    & $0.855 \pm 0.058$ & $0.924 \pm 0.019$ & $0.939 \pm 0.021$ \\
                               & FP rate & $2.95\%$          & $5.13\%$          & $6.41\%$ \\
\hline
\multirow{2}{*}{+ 3D synth}   & Dice    & $0.819 \pm 0.055$ & $0.912 \pm 0.035$ & $0.939 \pm 0.029$ \\
                               & FP rate & $2.18\%$          & $4.10\%$          & $6.54\%$ \\
\hline
\end{tabular}
\end{table}

\subsection{Comparison of Synthetic Augmentation Strategies}
Because we evaluate six pairwise comparisons across three data regimes and two metrics (36 tests in total), reporting uncorrected $p$-values would inflate the chance of spurious significance: at $\alpha = 0.05$, roughly one or two ``significant'' results would be expected by chance alone. We therefore control the false discovery rate using the Benjamini-Hochberg (BH) procedure~\cite{benjamini1995controlling}, which adjusts $p$-values to bound the expected proportion of false positives among the comparisons declared significant. All $p$-values reported below and in Table~\ref{tab:sig} are BH-adjusted over the full family of 36 tests.

After correction, a clear and consistent picture emerges. 2.5D and 3D augmentation both significantly outperform 2D on Dice and on false-positive rate in nearly every regime, and both significantly reduce the false-positive rate relative to real-only training at all three regimes (all BH-adjusted $p < 0.05$;  able~\ref{tab:sig}). Their effect on Dice is more regime-dependent: neither method significantly improves Dice over real-only at the extreme 2.5\% setting, but 2.5D does so at 5\% and 10\%, reflecting that specificity, rather than overlap, is the primary and most robust benefit of augmentation.

Critically, 2.5D and 3D (MAISI) are statistically indistinguishable from one another. No comparison between them reaches significance on any metric at any regime (BH-adjusted $p = 0.149$, $0.438$, $0.902$ for Dice; all $p > 0.39$ for FP rate), and their Dice scores converge to within a thousandth at 10\% ($0.938$ vs.\ $0.939$). Full 3D volumetric synthesis therefore confers no measurable advantage over the proposed 2.5D approach for this task.

These results suggest that the added complexity of full 3D volumetric synthesis (MAISI) does not translate into measurable performance gains over the proposed 2.5D approach. 2D synthesis, by contrast, is not a reliable augmentation strategy within given setup as it fails to improve, and can actively harm, segmentation performance when real data is scarce. The 2.5D method thus offers an effective, computationally lighter alternative to 3D synthesis for this task, achieving equivalent improvements in specificity and segmentation accuracy across all evaluated data regimes. 

\begin{table}[t]
\centering
\caption{Pairwise significance ($p$-values, paired $t$-test) for Dice between augmentation conditions, per regime.}
\label{tab:sig}
\begin{tabular}{lccc}
\hline
Comparison & 2.5\% real & 5\% real & 10\% real \\
\hline
Real vs.\ 2D     & $\mathbf{0.000}$ & $0.294$ & $0.840$ \\
Real vs.\ 2.5D   & $0.485$ & $\mathbf{0.005}$ & $\mathbf{0.006}$ \\
Real vs.\ 3D     & $0.221$ & $\mathbf{0.035}$ & $\mathbf{0.035}$ \\
2D vs.\ 2.5D     & $\mathbf{0.000}$ & $\mathbf{0.003}$ & $\mathbf{0.000}$ \\
2D vs.\ 3D       & $\mathbf{0.031}$ & $\mathbf{0.005}$ & $\mathbf{0.010}$ \\
2.5D vs.\ 3D     & $0.103$ & $0.365$ & $0.995$ \\
\hline
\end{tabular}
\end{table}

\begin{table}[t]
\centering
\caption{Pairwise BH-adjusted $p$-values (paired $t$-test) per regime.
Bold = $p < 0.05$; winner shown in parentheses where significant.
BH-adjusted over 36 tests (6 comparisons $\times$ 3 regimes $\times$ 2 metrics).}
\label{tab:sig_pvalues}
\setlength{\tabcolsep}{6pt}

\smallskip
\textit{(a) Dice score} ($\uparrow$ higher is better)

\smallskip
\begin{tabular}{lccc}
\hline
Comparison & 2.5\% & 5\% & 10\% \\
\hline
Real vs.\ +2D    & \textbf{.0004} (Real)  & .391        & .878        \\
Real vs.\ +2.5D  & .545                   & \textbf{.010} (2.5D) & \textbf{.011} (2.5D) \\
Real vs.\ +3D    & .306                   & .053        & .053        \\
+2D vs.\ +2.5D   & \textbf{$<$.001} (2.5D) & \textbf{.007} (2.5D) & \textbf{$<$.001} (2.5D) \\
+2D vs.\ +3D     & .051                   & \textbf{.011} (3D)  & \textbf{.018} (3D)  \\
+2.5D vs.\ +3D   & .149                   & .438        & .995        \\
\hline
\end{tabular}

\bigskip
\textit{(b) FP rate (\%)} ($\downarrow$ lower is better)

\smallskip
\begin{tabular}{lccc}
\hline
Comparison & 2.5\% & 5\% & 10\% \\
\hline
Real vs.\ +2D    & .675                    & \textbf{.002} (+2D)  & .393        \\
Real vs.\ +2.5D  & \textbf{$<$.001} (2.5D) & \textbf{$<$.001} (2.5D) & \textbf{.002} (2.5D) \\
Real vs.\ +3D    & \textbf{$<$.001} (3D)   & \textbf{$<$.001} (3D)   & \textbf{.001} (3D)   \\
+2D vs.\ +2.5D   & \textbf{$<$.001} (2.5D) & \textbf{.013} (2.5D) & \textbf{$<$.001} (2.5D) \\
+2D vs.\ +3D     & \textbf{$<$.001} (3D)   & \textbf{.008} (3D)   & \textbf{$<$.001} (3D)   \\
+2.5D vs.\ +3D   & .489                    & .393        & .878        \\
\hline
\end{tabular}
\end{table}

\subsection{Cross-Slice Consistency} \label{sec:consistency}
\begin{figure*}[ht]
  \centering
  \includegraphics[width=\textwidth]{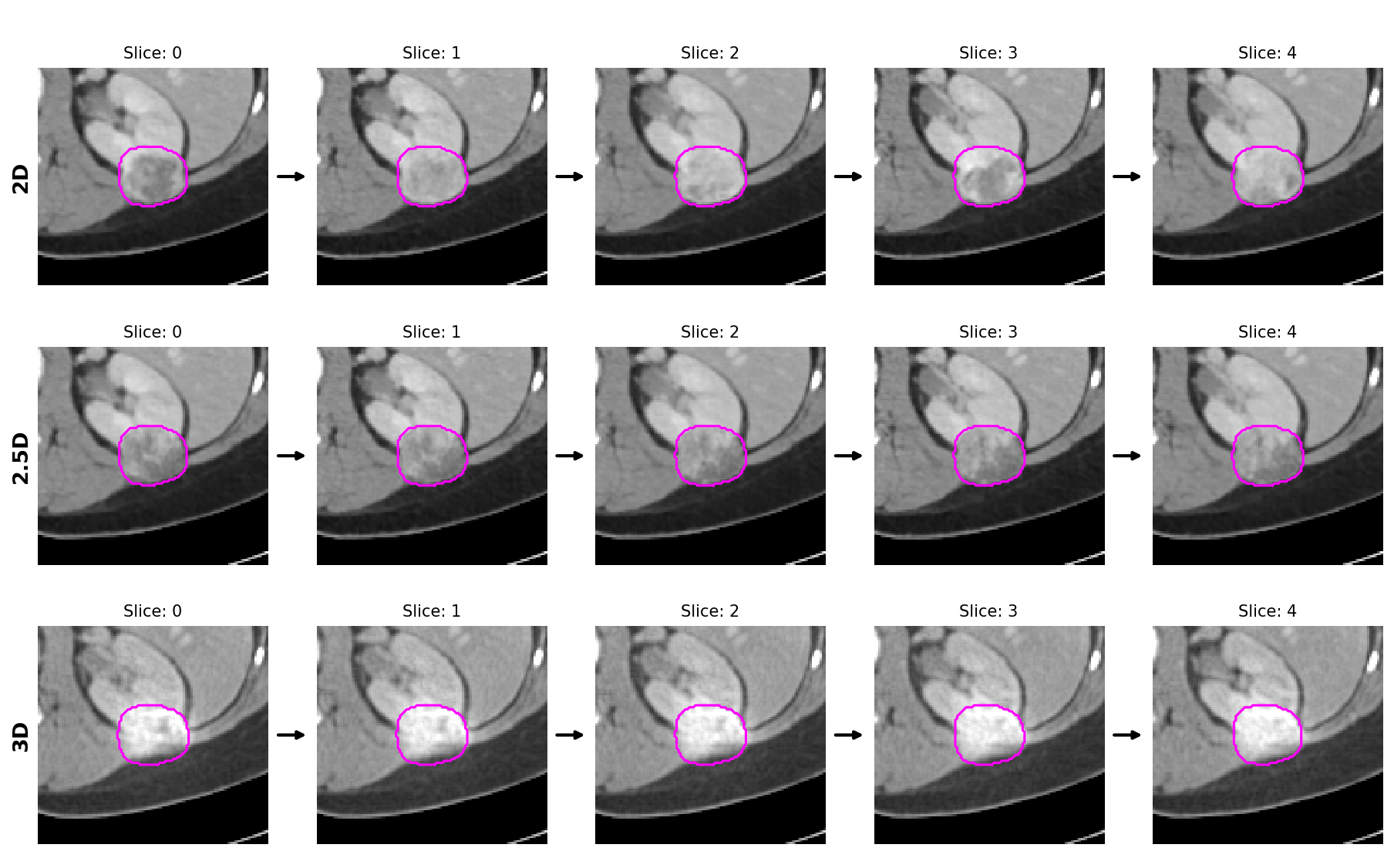}
  \caption{Tumor crops from five consecutive axial slices generated by each synthesis strategy. Each row shows the same lesion across adjacent slices; outlines indicate the tumor boundary. The 2D method exhibits pronounced changes in texture and contrast between consecutive slices, whereas the context-aware methods (2.5D, 3D) maintain a consistent tumor appearance throughout the volume.}
  \label{fig:slice_consistency}
\end{figure*}

To examine the mechanism underlying the poor performance of 2D augmentation, we visually compared the volumetric coherence of tumors generated by each strategy across consecutive axial slices (Figure~\ref{fig:slice_consistency}). The 2D method, which synthesizes each slice independently, produced tumors whose internal texture, contrast, and boundary shape varied noticeably from one slice to the next, yielding an incoherent appearance when viewed as a volume. In contrast, the 2.5D and 3D (MAISI) methods, both of which incorporate inter-slice context during generation, produced lesions that remained visually consistent across adjacent slices, with smoothly varying texture and stable boundaries in the $z$-direction.

This observation supports our interpretation that the failure of 2D augmentation is driven by a lack of volumetric coherence rather than poor per-slice realism: individual 2D slices appear locally plausible, but their inconsistency across the volume conflicts with the 3D structural priors learned by nnU-Net, introducing a training signal that degrades rather than improves segmentation under scarce real data.

To complement this qualitative assessment, we quantified cross-slice consistency by computing the mean absolute intensity difference between the tumor regions of adjacent slices, averaged over all synthetic lesions ($n = 160$ per method). The 2D method showed substantially higher inter-slice variation (mean adjacent-slice difference $22.25 \pm 5.52$) than the 2.5D ($6.02 \pm 2.53$) and 3D ($6.16 \pm 2.08$) strategies which results in roughly a $3.7\times$ larger jump in tumor intensity between consecutive slices. This confirms that the visual inconsistency of independently-synthesized 2D slices translates into a measurable lack of volumetric coherence, whereas the context-aware 2.5D and 3D methods produce tumors that vary smoothly through the volume.

\subsection{FID}
FID was computed on 2D axial tumor crops rather than on whole volumes. Because the synthetic scans are real patient images in which only a small tumor region has been modified, FID over full volumes would be dominated by the shared, unmodified anatomy and would be uninformative about tumor quality. Restricting the computation to $128 \times 128$ crops centred on the tumor in each tumor-bearing slice removes this shared background and isolates the generated region, at the cost of substantially higher absolute values than are typically reported for whole-image FID.

Against the real TUH cohort tumor distribution, 2D and 2.5D synthesis achieved comparable scores ($66.7$ and $66.0$), while 3D (MAISI) was worse ($80.8$); the same ordering held against the KiTS+KIRC training distribution ($77.1$, $76.2$, and $80.0$; Table~\ref{tab:fid}). The 2D and 2.5D methods were closer to the TUH dataset distribution than to the training distribution, consistent with their tumors being inpainted into real TUH control scans and therefore inheriting the acquisition characteristics of the target site. MAISI, which synthesizes the full volume rather than inpainting into existing anatomy, showed no such shift.

Critically, FID did not rank the methods in the same order as downstream segmentation performance. 2D matched 2.5D on both references despite performing substantially worse as augmentation, and 3D scored worst on FID while performing comparably to 2.5D in segmentation. Part of this is structural: FID here is computed on independent 2D crops using features from an ImageNet-pretrained Inception network, so it captures per-slice texture realism only and is by construction blind to the cross-slice inconsistency (Section~\ref{sec:consistency}) that our results identify as the property governing augmentation utility. More broadly, this misalignment is consistent with recent evidence that FID and related feature-distance metrics fail to predict downstream task performance in biomedical image synthesis, where the purpose of generation is to enrich training data rather than to produce perceptually convincing images; \cite{wu2025pragmatic} report the same dissociation across multiple generative models and retinal imaging modalities and recommend prioritizing downstream evaluation. Our findings support that recommendation in the context of volumetric CT tumor synthesis, therefore, we report FID as a supplementary indicator of local texture realism, while treating segmentation performance as the primary criterion for evaluating synthetic augmentation.

\begin{table}[t]
\centering
\caption{Tumor-region FID for each synthesis strategy against two references: the diffusion training distribution  (KiTS23+KIRC) and the real internal TUH cohort tumor distribution ($n=191$). FID computed on fixed-size axial tumor crops; lower is better. Values are relative indicators (ImageNet-pretrained Inception features on grayscale CT; $n_{\text{crops}}$ reported below).}

\label{tab:fid}
\begin{tabular}{lcc}
\hline
Method & FID vs.\ KiTS+KIRC & FID vs.\ TUH real \\
\hline
2D (ours)   & $77.13$ & $66.65$ \\
2.5D (ours) & $\mathbf{76.20}$ & $\mathbf{65.97}$ \\
3D (MAISI)  & $80.02$ & $80.75$ \\
\hline
\end{tabular}
\end{table}

\subsection{Discussion}

The results demonstrate that the choice of synthetic augmentation strategy has a decisive impact on model behavior under low-data conditions. The consistent failure of 2D augmentation to reduce false positives at 2.5\% real data, and its instability in FP object counts more broadly, suggests that slice-level synthesis produces tumors with inconsistent volumetric coherence, which may confuse the segmentation model when real training signal is scarce. 2.5D synthesis, which conditions on adjacent slices to enforce local 3D consistency, appears sufficient to overcome this limitation without requiring full volumetric synthesis.

The primary benefit of synthetic augmentation at all regimes is specificity rather than sensitivity: Dice scores are maintained or modestly improved, while false positives are reduced substantially. This is clinically relevant, as over-detection in control scans leads to unnecessary follow-up procedures. The FP suppression achieved by 2.5D and 3D augmentation --- reducing FP rates from $\sim$18--20\% to $\sim$3--6\% across regimes --- represents a practically meaningful improvement even where Dice differences are not statistically significant.

The near-equivalence of 2.5D and 3D (MAISI) across all metrics and regimes is a notable finding. MAISI is a large-scale, resource-intensive 3D diffusion model requiring substantial compute for both training and inference. The proposed 2.5D approach achieves indistinguishable results while operating at substantially lower computational cost, making it more accessible for clinical deployment settings where GPU resources are limited.

Finally, the high cross-split variability observed at 2.5\% real data warrants practical caution: with only 4 real tumor scans, the specific cases selected exert more influence on outcomes than any augmentation strategy. This underscores the importance of multi-split evaluation and suggests that 8 real tumor-positive cases (5\%) provides a more stable foundation for synthetic augmentation to deliver consistent benefits in our data setup.

\section{Conclusion}
We proposed a diffusion-based inpainting framework for renal tumor synthesis that generates anatomically plausible kidney tumors directly within healthy CT scans from a clinical partner site, requiring no additional annotation of the generated data. Across three low-data regimes (2.5\%, 5\%, 10\% of available tumor cases), we evaluated 2D, 2.5D, and 3D (MAISI) synthesis strategies as augmentation for nnU-Net segmentation.
 
The main finding is that 2.5D and 3D augmentation both substantially and consistently reduce false positive rates compared to real-data-only training with this effect significant at all regimes after. 2D augmentation, by contrast, provides no consistent benefit and actively hurts Dice performance at the most constrained setting. Crucially, 2.5D synthesis matches full 3D synthesis (MAISI) on all metrics at all regimes, despite operating at considerably lower computational cost.
 
These results suggest that, for this task, local volumetric consistency during synthesis, enforced by conditioning on adjacent slices, is sufficient to produce useful training data, without the complexity and resource requirements of full 3D generation. The proposed framework is a practical option for clinical settings where annotated tumor data is scarce and GPU resources are limited.

\section{Acknowledgments}
The project was supported by European Union and Estonian Research Council via the project TEM-TA101. The computational resources were provided by the High Performance Computing Cluster at the University of Tartu. We would like to thank Tartu University Hospital and Better Medicine for their support and data provision.
%
%
\bibliographystyle{unsrt} 

\bibliography{mybibliography}

@article{goceri2023medical,
  title={Medical image data augmentation: techniques, comparisons and interpretations},
  author={Goceri, Evgin},
  journal={Artificial intelligence review},
  volume={56},
  number={11},
  pages={12561--12605},
  year={2023},
  publisher={Springer}
}

@article{foroozandeh2020synthesizing,
  title={Synthesizing brain tumor images and annotations by combining progressive growing GAN and SPADE},
  author={Foroozandeh, Mehdi and Eklund, Anders},
  journal={arXiv preprint arXiv:2009.05946},
  year={2020}
}

@inproceedings{chen2024towards,
  title={Towards generalizable tumor synthesis},
  author={Chen, Qi and Chen, Xiaoxi and Song, Haorui and Xiong, Zhiwei and Yuille, Alan and Wei, Chen and Zhou, Zongwei},
  booktitle={Proceedings of the IEEE/CVF conference on computer vision and pattern recognition},
  pages={11147--11158},
  year={2024}
}

@article{jin2021free, 
  title={Free-form tumor synthesis in computed tomography images via richer generative adversarial network},
  author={Jin, Qiangguo and Cui, Hui and Sun, Changming and Meng, Zhaopeng and Su, Ran},
  journal={Knowledge-Based Systems},
  volume={218},
  pages={106753},
  year={2021},
  publisher={Elsevier}
}

@article{han2019combining,
  title={Combining noise-to-image and image-to-image GANs: Brain MR image augmentation for tumor detection},
  author={Han, Changhee and Rundo, Leonardo and Araki, Ryosuke and Nagano, Yudai and Furukawa, Yujiro and Mauri, Giancarlo and Nakayama, Hideki and Hayashi, Hideaki},
  journal={Ieee Access},
  volume={7},
  pages={156966--156977},
  year={2019},
  publisher={IEEE}
}

@article{ho2020denoising,
  title={Denoising diffusion probabilistic models},
  author={Ho, Jonathan and Jain, Ajay and Abbeel, Pieter},
  journal={Advances in neural information processing systems},
  volume={33},
  pages={6840--6851},
  year={2020}
}

@inproceedings{pinaya2022brain,
  title={Brain imaging generation with latent diffusion models},
  author={Pinaya, Walter HL and Tudosiu, Petru-Daniel and Dafflon, Jessica and Da Costa, Pedro F and Fernandez, Virginia and Nachev, Parashkev and Ourselin, Sebastien and Cardoso, M Jorge},
  booktitle={MICCAI workshop on deep generative models},
  pages={117--126},
  year={2022},
  organization={Springer}
}

@inproceedings{saharia2022palette,
  title={Palette: Image-to-image diffusion models},
  author={Saharia, Chitwan and Chan, William and Chang, Huiwen and Lee, Chris and Ho, Jonathan and Salimans, Tim and Fleet, David and Norouzi, Mohammad},
  booktitle={ACM SIGGRAPH 2022 conference proceedings},
  pages={1--10},
  year={2022}
}

@misc{Akin2016_TCGA_KIRC,
  author={Akin, O. and Elnajjar, P. and Heller, M. and Jarosz, R. and Erickson, B. J. and Kirk, S. and Lee, Y. and Linehan, M. W. and Gautam, R. and Vikram, R. and Garcia, K. M. and Roche, C. and Bonaccio, E. and Filippini, J.},
  title={The Cancer Genome Atlas Kidney Renal Clear Cell Carcinoma Collection (TCGA-KIRC)},
  year={2016},
  version={3},
  publisher={The Cancer Imaging Archive},
  doi={10.7937/K9/TCIA.2016.V6PBVTDR},
  url={https://doi.org/10.7937/K9/TCIA.2016.V6PBVTDR}
}

@misc{heller2023kits21,
      title={The KiTS21 Challenge: Automatic segmentation of kidneys, renal tumors, and renal cysts in corticomedullary-phase CT}, 
      author={Nicholas Heller and Fabian Isensee and Dasha Trofimova and Resha Tejpaul and Zhongchen Zhao and Huai Chen and Lisheng Wang and Alex Golts and Daniel Khapun and Daniel Shats and Yoel Shoshan and Flora Gilboa-Solomon and Yasmeen George and Xi Yang and Jianpeng Zhang and Jing Zhang and Yong Xia and Mengran Wu and Zhiyang Liu and Ed Walczak and Sean McSweeney and Ranveer Vasdev and Chris Hornung and Rafat Solaiman and Jamee Schoephoerster and Bailey Abernathy and David Wu and Safa Abdulkadir and Ben Byun and Justice Spriggs and Griffin Struyk and Alexandra Austin and Ben Simpson and Michael Hagstrom and Sierra Virnig and John French and Nitin Venkatesh and Sarah Chan and Keenan Moore and Anna Jacobsen and Susan Austin and Mark Austin and Subodh Regmi and Nikolaos Papanikolopoulos and Christopher Weight},
      year={2023},
      eprint={2307.01984},
      archivePrefix={arXiv},
      primaryClass={cs.CV}
}

@article{isensee2021nnu,
  title={nnU-Net: a self-configuring method for deep learning-based biomedical image segmentation},
  author={Isensee, Fabian and Jaeger, Paul F and Kohl, Simon AA and Petersen, Jens and Maier-Hein, Klaus H},
  journal={Nature methods},
  volume={18},
  number={2},
  pages={203--211},
  year={2021},
  publisher={Nature Publishing Group}
}

@article{dorjsembe2024conditional,
  title={Conditional diffusion models for semantic 3D brain MRI synthesis},
  author={Dorjsembe, Zolnamar and Pao, Hsing-Kuo and Odonchimed, Sodtavilan and Xiao, Furen},
  journal={IEEE Journal of Biomedical and Health Informatics},
  volume={28},
  number={7},
  pages={4084--4093},
  year={2024},
  publisher={IEEE}
}

@article{huijben2024denoising,
  title={Denoising diffusion probabilistic models for addressing data limitations in chest X-ray classification},
  author={Huijben, Evi MC and Pluim, Josien PW and van Eijnatten, Maureen AJM},
  journal={Informatics in Medicine Unlocked},
  volume={50},
  pages={101575},
  year={2024},
  publisher={Elsevier}
}

@inproceedings{moghadam2023morphology,
  title={A morphology focused diffusion probabilistic model for synthesis of histopathology images},
  author={Moghadam, Puria Azadi and Van Dalen, Sanne and Martin, Karina C and Lennerz, Jochen and Yip, Stephen and Farahani, Hossein and Bashashati, Ali},
  booktitle={Proceedings of the IEEE/CVF winter conference on applications of computer vision},
  pages={2000--2009},
  year={2023}
}

@inproceedings{shin2018medical,
  title={Medical image synthesis for data augmentation and anonymization using generative adversarial networks},
  author={Shin, Hoo-Chang and Tenenholtz, Neil A and Rogers, Jameson K and Schwarz, Christopher G and Senjem, Matthew L and Gunter, Jeffrey L and Andriole, Katherine P and Michalski, Mark},
  booktitle={International workshop on simulation and synthesis in medical imaging},
  pages={1--11},
  year={2018},
  organization={Springer}
}

@article{luo2025review,
  title={Review of diffusion models and its applications in biomedical informatics},
  author={Luo, Jiawei and Yang, Liren and Liu, Yan and Hu, Changbao and Wang, Grant and Yang, Yan and Yang, Tie-Lin and Zhou, Xiaobo},
  journal={BMC Medical Informatics and Decision Making},
  volume={25},
  number={1},
  pages={390},
  year={2025},
  publisher={Springer}
}

@article{nalepa2019data,
  title={Data augmentation for brain-tumor segmentation: a review},
  author={Nalepa, Jakub and Marcinkiewicz, Michal and Kawulok, Michal},
  journal={Frontiers in computational neuroscience},
  volume={13},
  pages={83},
  year={2019},
  publisher={Frontiers Media SA}
}

@inproceedings{lugmayr2022repaint,
  title={Repaint: Inpainting using denoising diffusion probabilistic models},
  author={Lugmayr, Andreas and Danelljan, Martin and Romero, Andres and Yu, Fisher and Timofte, Radu and Van Gool, Luc},
  booktitle={Proceedings of the IEEE/CVF conference on computer vision and pattern recognition},
  pages={11461--11471},
  year={2022}
}

@article{song2020score,
  title={Score-based generative modeling through stochastic differential equations},
  author={Song, Yang and Sohl-Dickstein, Jascha and Kingma, Diederik P and Kumar, Abhishek and Ermon, Stefano and Poole, Ben},
  journal={arXiv preprint arXiv:2011.13456},
  year={2020}
}

@inproceedings{wolleb2022diffusion,
  title={Diffusion models for medical anomaly detection},
  author={Wolleb, Julia and Bieder, Florentin and Sandk{\"u}hler, Robin and Cattin, Philippe C},
  booktitle={International Conference on Medical image computing and computer-assisted intervention},
  pages={35--45},
  year={2022},
  organization={Springer}
}

@article{yi2019generative,
  title={Generative adversarial network in medical imaging: A review},
  author={Yi, Xin and Walia, Ekta and Babyn, Paul},
  journal={Medical image analysis},
  volume={58},
  pages={101552},
  year={2019},
  publisher={Elsevier}
}

@article{rais2024exploring,
  title={Exploring variational autoencoders for medical image generation: a comprehensive study},
  author={Rais, Khadija and Amroune, Mohamed and Benmachiche, Abdelmadjid and Haouam, Mohamed Yassine},
  journal={arXiv preprint arXiv:2411.07348},
  year={2024}
}

@article{kingma2013auto,
  title={Auto-encoding variational bayes},
  author={Kingma, Diederik P and Welling, Max},
  journal={arXiv preprint arXiv:1312.6114},
  year={2013}
}

@article{wasserthal2023totalsegmentator,
  title={TotalSegmentator: robust segmentation of 104 anatomic structures in CT images},
  author={Wasserthal, Jakob and Breit, Hanns-Christian and Meyer, Manfred T and Pradella, Maurice and Hinck, Daniel and Sauter, Alexander W and Heye, Tobias and Boll, Daniel T and Cyriac, Joshy and Yang, Shan and others},
  journal={Radiology: Artificial Intelligence},
  volume={5},
  number={5},
  pages={e230024},
  year={2023},
  publisher={Radiological Society of North America}
}

@inproceedings{guo2025maisi,
  title={Maisi: Medical ai for synthetic imaging},
  author={Guo, Pengfei and Zhao, Can and Yang, Dong and Xu, Ziyue and Nath, Vishwesh and Tang, Yucheng and Simon, Benjamin and Belue, Mason and Harmon, Stephanie and Turkbey, Baris and others},
  booktitle={2025 IEEE/CVF Winter Conference on Applications of Computer Vision (WACV)},
  pages={4430--4441},
  year={2025},
  organization={IEEE}
}

@inproceedings{rombach2022high,
  title={High-resolution image synthesis with latent diffusion models},
  author={Rombach, Robin and Blattmann, Andreas and Lorenz, Dominik and Esser, Patrick and Ommer, Bj{\"o}rn},
  booktitle={Proceedings of the IEEE/CVF conference on computer vision and pattern recognition},
  pages={10684--10695},
  year={2022}
}

@article{lepcha2025deep,
  title={Deep Learning in Medical Image Analysis: A Comprehensive Review of Algorithms, Trends, Applications, and Challenges},
  author={Lepcha, Dawa Chyophel and Goyal, Bhawna and Dogra, Ayush and Alkhayyat, Ahmed and Sahu, Prabhat Kumar and Ali, Aaliya and Kukreja, Vinay},
  journal={Computer Modeling in Engineering \& Sciences},
  volume={145},
  number={2},
  pages={1487},
  year={2025},
  publisher={Tech Science Press}
}

@article{heller2021state,
  title={The state of the art in kidney and kidney tumor segmentation in contrast-enhanced CT imaging: Results of the KiTS19 challenge},
  author={Heller, Nicholas and Isensee, Fabian and Maier-Hein, Klaus H and Hou, Xiaoshuai and Xie, Chunmei and Li, Fengyi and Nan, Yang and Mu, Guangrui and Lin, Zhiyong and Han, Miofei and others},
  journal={Medical image analysis},
  volume={67},
  pages={101821},
  year={2021},
  publisher={Elsevier}
}

@article{kebaili2023deep,
  title={Deep learning approaches for data augmentation in medical imaging: a review},
  author={Kebaili, Aghiles and Lapuyade-Lahorgue, J{\'e}r{\^o}me and Ruan, Su},
  journal={Journal of imaging},
  volume={9},
  number={4},
  pages={81},
  year={2023},
  publisher={MDPI}
}

@article{wang2024comprehensive,
  title={A comprehensive survey on deep active learning in medical image analysis},
  author={Wang, Haoran and Jin, Qiuye and Li, Shiman and Liu, Siyu and Wang, Manning and Song, Zhijian},
  journal={Medical Image Analysis},
  volume={95},
  pages={103201},
  year={2024},
  publisher={Elsevier}
}

@article{saad2024survey,
  title={A survey on training challenges in generative adversarial networks for biomedical image analysis},
  author={Saad, Muhammad Muneeb and O’Reilly, Ruairi and Rehmani, Mubashir Husain},
  journal={Artificial Intelligence Review},
  volume={57},
  number={2},
  pages={19},
  year={2024},
  publisher={Springer}
}

@article{heusel2017gans,
  title={Gans trained by a two time-scale update rule converge to a local nash equilibrium},
  author={Heusel, Martin and Ramsauer, Hubert and Unterthiner, Thomas and Nessler, Bernhard and Hochreiter, Sepp},
  journal={Advances in neural information processing systems},
  volume={30},
  year={2017}
}

@article{benjamini1995controlling,
  title={Controlling the false discovery rate: a practical and powerful
         approach to multiple testing},
  author={Benjamini, Yoav and Hochberg, Yosef},
  journal={Journal of the Royal Statistical Society: Series B
           (Methodological)},
  volume={57}, number={1}, pages={289--300}, year={1995},
  publisher={Wiley}
}

@article{wu2025pragmatic,
  title={A Pragmatic Note on Evaluating Generative Models with Fr$\backslash$'echet Inception Distance for Retinal Image Synthesis},
  author={Wu, Yuli and Liu, Fucheng and Yilmaz, R{\"u}veyda and Konermann, Henning and Walter, Peter and Stegmaier, Johannes},
  journal={arXiv preprint arXiv:2502.17160},
  year={2025}
}

\end{document}